\documentclass[11pt]{article}

\usepackage[preprint]{acl}

\usepackage{times}
\usepackage{latexsym}
\usepackage{tcolorbox}
\usepackage{enumitem}   
\usepackage{pifont}

\usepackage[T1]{fontenc}

\usepackage[utf8]{inputenc}

\usepackage{microtype}

\usepackage{inconsolata}

\usepackage{graphicx}
\usepackage{adjustbox}  

\usepackage{amsmath}
\usepackage{amssymb}

\usepackage{float}
\usepackage{algorithm}
\usepackage{algorithmic}

\usepackage{enumitem}
\usepackage{wasysym}  

\usepackage{tikz}
\usetikzlibrary{arrows.meta, positioning, calc}

\usepackage{booktabs}
\usepackage{multirow}

\newcommand{\benchmarkname}{AD-Bench}

\usepackage{pdfpages}

\title{\benchmarkname{}: A Real-World, Trajectory-Aware Advertising Analytics Benchmark for LLM Agents}

\author{
\textbf{Lingxiang Hu}$^{*}$, \textbf{Yiding Sun}$^{*}$, \textbf{Tianle Xia}$^{*}$, \textbf{Wenwei Li}$^{*}$, \textbf{Ming Xu}$^{*\dagger}$ \\
\textbf{Lan Xu}, \textbf{Siying Wang}, \textbf{Wei Xu}, \textbf{Jie Jiang} \\
Tencent \\
\texttt{\{lingxianghu,emanuelsun,tianlexia,wenweiwwli,flemingxu\}@tencent.com} \\
\texttt{\{lanxu,siyingwang,davidxu,zeus\}@tencent.com} \\
{\small $^{*}$Equal contribution. \quad $^{\dagger}$Corresponding author.}
}

\begin{document}
\maketitle
\begin{abstract}

While Large Language Model (LLM) agents have made remarkable progress on complex reasoning, evaluating them in real-world environments remains an open problem. Existing benchmarks are largely confined to idealized simulations and fail to capture specialized domains such as advertising and marketing analytics, where tasks require multi-round interaction with professional tools and where ground-truth answers quickly become obsolete as data and platform rules evolve.

To address this, we propose \textbf{\benchmarkname{}}, a benchmark built from real user marketing-analysis requests on a production advertising platform. \benchmarkname{} introduces two key designs: (i) a \emph{dynamic ground-truth pipeline} that replays expert tool-call trajectories to regenerate answers consistent with the current environment, mitigating answer obsolescence; and (ii) a \emph{trajectory-aware evaluation} that jointly measures end-to-end answer correctness (Pass@$k$) and trajectory coverage. Requests are stratified into three difficulty levels (L1--L3) to probe multi-round, multi-tool collaboration. Experiments show that the best model, Claude-Opus-4.7, attains 
Pass@1 = 76.9\% and Pass@3 = 80.4\% with 82.7\% trajectory coverage overall, yet drops sharply on L3 to Pass@1 = 61.4\% and Pass@3 = 65.1\%, revealing that even state-of-the-art agents have substantial gaps in complex advertising analytics. 


\end{abstract}

\section{Introduction}
\label{sec:introduction}

LLMs evolve from passive knowledge-retrieval interfaces into autonomous agents that can reason, plan, and take actions in real-world environments~\cite{wang2024llmagentsurvey,xi2025llmagentsurvey,kimi-k2,zeng2025futurex,xia2025lact}. Unlike conventional question answering~(QA), real-world agents must interact with tools through multiple steps, including retrieval-augmented generation (RAG)~\cite{lewis2020rag} and access to user-private data, ultimately producing effective answers.

However, most existing agent evaluation benchmarks are built on fixed tasks in idealized or simulated settings~\cite{mialon2023gaia,browsecomp,zhou2024webarena,xbench2025,wong2025widesearch,yao2024taubench,tau2bench,bfcl}, and thus struggle to capture the dynamics and complexity of real-world production settings arising from continuously evolving environment states and multi-tool interactions. This limitation is particularly pronounced in advertising and marketing, manifesting in the following three challenges:



\noindent\textbullet\quad \textbf{Challenge 1: Ground-Truth Obsolescence.} This challenge is particularly evident in advertising and marketing, where user data, campaign delivery conditions, and platform policies are frequently updated. Consequently, benchmarks based on static QA pairs can quickly become obsolete: once the underlying data or governing rules change, previously correct answers may no longer remain valid, compromising the reliability of the ground truth.

\noindent\textbullet\quad \textbf{Challenge 2: Lack of Evaluation for Advertising Trajectories.} In advertising analytics, user queries often require an agent to plan and execute an execution trajectory over a complex set of tools. This involves selecting appropriate tools and skills~\cite{grasp2026,skillsbench2026} from a diverse toolkit and specifying the correct parameters at each step. Although end-to-end evaluation can measure the accuracy of the final answer, it is insufficient for assessing the validity of the agent’s execution trajectory in advertising and marketing scenarios.

\noindent\textbullet\quad \textbf{Challenge 3: Long-Tail Distribution of Advertising Analytics.} Customer analytics needs served by marketing platforms vary substantially. In real online traffic, the frequency of analytics tasks typically follows a long-tail distribution: basic descriptive analyses account for the majority of requests, while highly specialized needs are dispersed across the long tail. If evaluation covers only high-frequency tasks, it can introduce selection bias and thus overestimate the agent’s overall capability in real-world settings.

To address the challenges above, we introduce \benchmarkname{}, a benchmark designed to evaluate agents on real-world advertising and marketing platforms in terms of trajectory coverage and end-to-end task success. Our main contributions are as follows:

\noindent\textbullet\quad \textbf{Dynamic Ground-Truth Generation Pipeline.}
To mitigate ground-truth obsolescence, when constructing our benchmark based on online analytical requests, we do not directly collect final answers from human experts. Instead, we record their problem-solving trajectories and associated tool calls. During evaluation, we replay these trajectories and re-execute the corresponding tool calls to regenerate ground-truth answers consistent with the current environment, thereby reducing errors caused by outdated answers.

\noindent\textbullet\quad \textbf{Trajectory-Aware Evaluation.}
We decompose evaluation into two complementary aspects: (i) end-to-end answer correctness on analytical requests, which directly measures agent reliability; and (ii) trajectory coverage, i.e., whether the expert reference trajectory is covered by the agent’s execution, providing finer-grained analysis.

\noindent\textbullet\quad \textbf{Difficulty-Stratified Task Classification.}
To counter the long-tail distribution of analytics needs, we organize the benchmark along a difficulty hierarchy instead of sampling only high-frequency requests. Using the complexity of expert trajectories as a proxy for task difficulty, we stratify analytical requests into three levels (L1--L3), ranging from single-metric retrieval to multi-condition synthesis and judgment. This stratification mitigates selection bias toward simple queries and enables a fine-grained assessment of agent capability across the full difficulty spectrum.

\begin{figure}[t]
    \centering
    \includegraphics[width=\columnwidth]{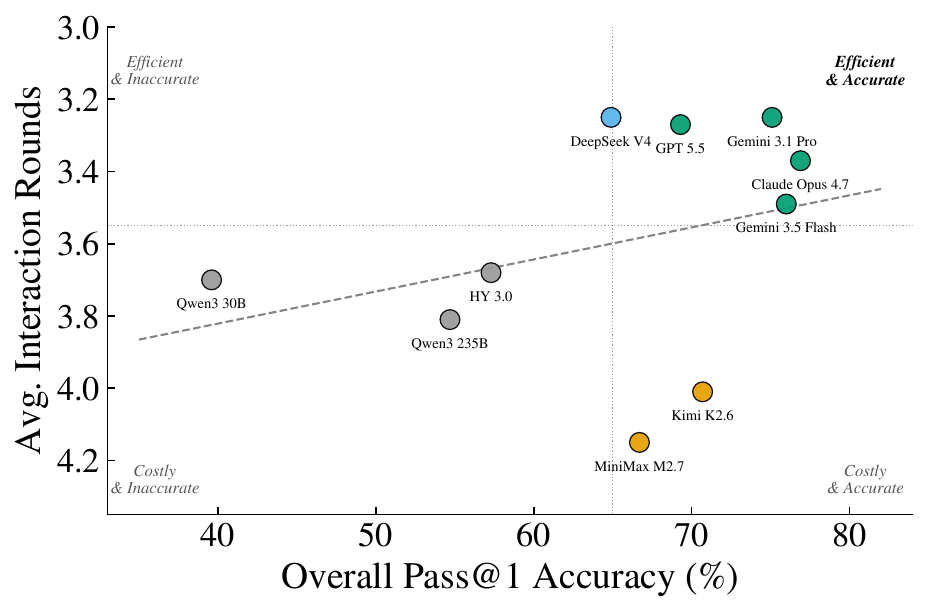}
    \caption{Efficiency--accuracy trade-off across evaluated models, plotting overall Pass@1 against the average number of interaction steps (fewer is better). Claude-Opus-4.7 attains the best accuracy (76.9\% Pass@1), yet no model reaches high accuracy with consistently short trajectories, underscoring the difficulty of \benchmarkname{}.}
    \label{fig:intro_l3}
\end{figure}

Experimental results in Section~\ref{sec:experiments} reveal a clear efficiency--accuracy trade-off across models (Figure~\ref{fig:intro_l3}): even the best overall model, Claude-Opus-4.7, attains only 76.9\% Pass@1, and its performance degrades sharply on the hardest (L3) queries, dropping to a Pass@3 of 65.1\% and a trajectory coverage of 69.9\%. Our error analysis spans six dimensions: tool errors, parameter errors, instruction-following errors, computation errors, hallucinations, and long-context limitations. Consequently, \benchmarkname{} establishes a more rigorous, sustainable, and business-aligned evaluation standard to facilitate the development of next-generation action-oriented marketing agents.

\section{\benchmarkname{}}
\label{sec:benchmark_construction}

\begin{figure*}[htbp]
    \centering
    \includegraphics[width=0.95\textwidth]{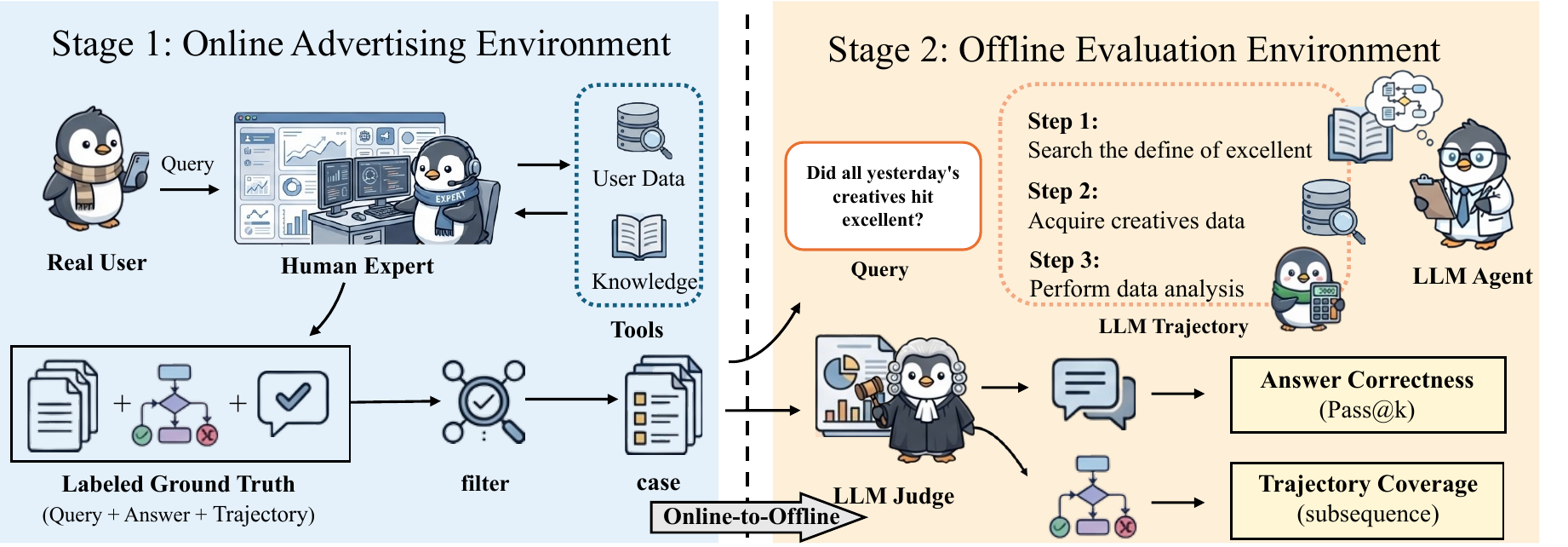}
    \caption{System overview of \textbf{\benchmarkname{}}. The online advertising environment (left) yields human-validated ground-truth trajectories, while the offline evaluation environment (right) evaluates LLM agents and uses LLM-as-a-judge to assess answer correctness and trajectory coverage.}
    \label{fig:system_overview}
\end{figure*}

\subsection{Benchmark Construction}
The construction of \benchmarkname{} relies on a real-world Online Advertising Environment (Figure~\ref{fig:system_overview}, left). We collect 2,000 native user analysis requests, and human experts resolve each one with standard marketing tools, recording their tool-call trajectories as $(\textup{Question}, \textup{Execution Trajectory})$ pairs. Rather than freezing static answers, the reference answer is regenerated on demand by replaying the trajectory against the current environment, keeping it correct under evolving business rules. After cleaning and deduplication, 823 high-quality instances remain, which are further curated into a final evaluation set of 225 instances (Section~\ref{subsec:task_curation}).

\subsection{Evaluation Framework} Figure~\ref{fig:system_overview}~(right) illustrates the evaluation pipeline of \benchmarkname{}. In the offline environment, a ReAct-based agent processes benchmark requests, simulating expert reasoning and tool calls. The execution follows three steps: retrieving business definitions for semantic alignment, fetching advertising metrics, and conducting analysis to generate a response; a detailed L3 case study is provided in Appendix Figure~\ref{fig:case_overview}. Finally, an LLM-judge evaluates the generated trajectory and response in terms of answer correctness and trajectory coverage.

\subsection{Task Classification and Reference-Trajectory Curation}
\label{subsec:task_curation}
To prevent the evaluation from being dominated by high-frequency simple queries while ensuring coverage of long-tail analytical needs, we stratified user requests. Specifically, we used the complexity of human expert trajectories as a proxy for task difficulty and categorized all requests into three levels. Through rigorous expert verification, for each of the 225 retained instances we designate an \emph{optimal reference trajectory}, i.e., the shortest, most direct expert solution path used to anchor trajectory coverage, while Pass@$k$ still credits correct answers reached via alternative or self-corrected paths. Table~\ref{tab:level_dist} summarizes the count, definition, and representative examples for each difficulty level (L1, L2, and L3).

\begin{table}[t]
\centering
\small
\resizebox{0.48\textwidth}{!}{%
\begin{tabular}{p{0.5cm}p{0.5cm} p{2.4cm} p{3.0cm}}
\toprule
\textbf{Level} & \textbf{Count} & \textbf{Definition} & \textbf{Example Query} \\
\midrule
L1 & 31 & Retrieve a metric for a fixed scope &
\textit{``Get the total cost across all accounts yesterday.''} \\
\addlinespace[2pt]
L2 & 111 & Compute metrics with filtering or comparison &
\textit{``Compute the week-over-week cost growth rate.''} \\
\addlinespace[2pt]
L3 & 83 & Synthesize multiple conditions and make a judgment &
\textit{``Analyze whether video and image CTRs both exceed the excellence threshold.''} \\
\bottomrule
\end{tabular}
}
\caption{Count, definition, and a representative example for each difficulty level in \benchmarkname{}.}
\label{tab:level_dist}
\end{table}

\subsection{Evaluation Metrics}
We evaluate agents from two complementary and mutually non-implied angles: \emph{whether} they reach the correct answer and \emph{how} they get there.
\begin{itemize}[leftmargin=*, noitemsep]
    \item \textbf{Answer correctness:} We report Pass@$k$ on the final answer. Pass@1 measures single-run correctness, while Pass@3 measures whether at least one of three independent runs is correct, accounting for sampling stochasticity.
    
    \item \textbf{Trajectory coverage:} A sample is \emph{covered} only when the agent reproduces the expert reference trajectory, i.e., it calls the required tools in a valid order, supplies the key parameters correctly, and follows the instruction constraints---no tool, parameter, or instruction-following error (Section~\ref{sec:error_analysis}). As some tools have a vast parameter space (e.g., RAG queries or code), we verify the \emph{key} parameters rather than exact matches.

\end{itemize}
\section{Experiments and Results}
\label{sec:experiments}

\begin{table*}[htbp]
    \centering
    \caption{Pass@1, Pass@3, and average trajectory length across overall and all difficulty tiers on \textbf{\benchmarkname{}}.}
    \label{tab:model_performance}
    \resizebox{\textwidth}{!}{%
    \begin{tabular}{lccccccccc}
        \toprule
        \multirow{2}{*}{\textbf{Model}} & \multicolumn{2}{c}{\textbf{Overall}} & \multicolumn{2}{c}{\textbf{L1}} & \multicolumn{2}{c}{\textbf{L2}} & \multicolumn{2}{c}{\textbf{L3}} & \multirow{2}{*}{\textbf{Avg.\ Len.}} \\
        \cmidrule(lr){2-3}\cmidrule(lr){4-5}\cmidrule(lr){6-7}\cmidrule(lr){8-9}
        & Pass@1$\uparrow$ & Pass@3$\uparrow$ & Pass@1$\uparrow$ & Pass@3$\uparrow$ & Pass@1$\uparrow$ & Pass@3$\uparrow$ & Pass@1$\uparrow$ & Pass@3$\uparrow$ & \\
        \midrule
        Claude-Opus-4.7 & \textbf{76.9} & \textbf{80.4} & \textbf{96.8} & \textbf{100.0} & \textbf{82.9} & \textbf{86.5} & \underline{61.4} & 65.1 & 3.37 \\
        Gemini-3.5-Flash & \underline{76.0} & \underline{79.6} & \underline{93.5} & 93.5 & \underline{82.0} & \underline{84.7} & \underline{61.4} & 67.5 & 3.49 \\
        Gemini-3.1-Pro & 75.1 & 78.7 & \textbf{96.8} & \textbf{100.0} & 77.5 & 79.3 & \textbf{63.9} & \underline{69.9} & \textbf{3.25} \\
        DeepSeek-V4 & 64.9 & 77.3 & \textbf{96.8} & \textbf{100.0} & 63.1 & 73.9 & 55.4 & \textbf{73.5} & \textbf{3.25} \\
        GPT-5.5 & 69.3 & 77.3 & 90.3 & 93.5 & 74.8 & 81.1 & 54.2 & 66.3 & \underline{3.27} \\
        Kimi-K2.6 & 70.7 & 76.9 & 90.3 & \underline{96.8} & \underline{82.0} & 82.9 & 48.2 & 61.4 & 4.01 \\
        MiniMax-M2.7 & 66.7 & 74.7 & 90.3 & 93.5 & 77.5 & 82.0 & 43.4 & 57.8 & 4.15 \\
        HY-3.0 & 57.3 & 68.0 & 64.5 & 74.2 & 71.2 & 79.3 & 36.1 & 50.6 & 3.68 \\
        \midrule
        Qwen3-235B & 54.7 & 64.0 & 90.3 & \underline{96.8} & 62.2 & 70.3 & 31.3 & 43.4 & 3.81 \\
        Qwen3-30B-A3B & 39.6 & 49.3 & 80.6 & 83.9 & 44.1 & 55.9 & 18.1 & 27.7 & 3.70 \\
        \bottomrule
    \end{tabular}%
    }
\end{table*}

\subsection{Experimental Setup}

We evaluate a set of open-weight and proprietary state-of-the-art LLMs. The open-weight models include Qwen3-235B and Qwen3-30B-A3B~\cite{qwen3}, DeepSeek-V4~\cite{deepseek2026v4}, Kimi-K2.6~\cite{kimi-k2}, MiniMax-M2.7~\cite{minimax2026m27}, and HY-3.0~\cite{tencent2026hy3}. The proprietary models are accessed via public APIs, including Claude-Opus-4.7~\cite{anthropic2026opus47}, Gemini-3.1-Pro~\cite{google2026gemini31pro}, Gemini-3.5-Flash~\cite{google2026gemini35flash}, and GPT-5.5~\cite{openai2026gpt55}. We implement a marketing agent based on the widely adopted ReAct framework~\cite{react} and equip it with 9 domain-specific tools for executing marketing tasks. We follow the evaluation protocol defined in Section~\ref{sec:benchmark_construction} and use Gemini-3.1-Pro as the judge to determine answer correctness and assess trajectory coverage. Additional experimental settings and results are provided in Appendix~\ref{sec:appendix_prompts}.

\subsection{Main Results and Analysis}

\textbf{Performance across difficulty tiers.} Table~\ref{tab:model_performance} reports the performance of all evaluated models on \benchmarkname{}. Overall, proprietary models set the current performance upper bound: Claude-Opus-4.7 achieves the best Pass@3 (80.4\%), followed closely by Gemini-3.5-Flash (79.6\%) and Gemini-3.1-Pro (78.7\%). However, aggregated scores mask a pronounced stratification by difficulty. While top models are highly reliable on L1 (Pass@3 > 93\%), performance drops sharply by 25--35 percentage points from L1 to L3 on tasks requiring complex, multi-tool orchestration. This indicates that current LLM agents are strong at direct information retrieval, yet remain fragile on the long-horizon dependency chains characteristic of industrial advertising analytics. Beyond offline evaluation, we further use \benchmarkname{} as a continuous evaluation harness to diagnose and iterate a production agent, raising its online user satisfaction rate to $88.37\%$ (Appendix~\ref{sec:deployment}).

\begin{figure*}[t]
    \centering
    \includegraphics[width=0.9\textwidth]{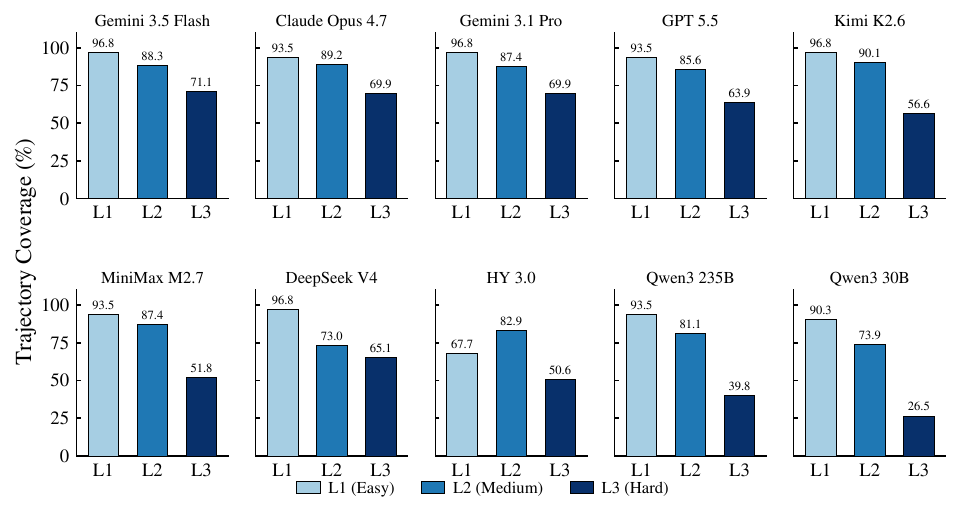}
    \caption{Trajectory coverage of each evaluated model across the three difficulty levels (L1/L2/L3). Coverage is near-saturated on L1 for most models but degrades substantially on L3, revealing reduced execution stability on complex tasks.}
    \label{fig:level_traj_coverage}
\end{figure*}

\textbf{Trajectory coverage and execution robustness.} To diagnose whether failures stem from tool-use planning or answer generation, we examine trajectory coverage across tiers. As shown in Figure~\ref{fig:level_traj_coverage}, most models achieve near-saturated coverage on L1 (>93\%), suggesting they can reliably identify the appropriate tools and parameters for low-complexity queries. By contrast, on L3 even the best overall model, Claude-Opus-4.7, sees its coverage drop to 69.9\%: the model increasingly deviates from the reference trajectory on complex tasks, reflecting failures in planning and executing the required tool-use steps.


\begin{figure}[t]
    \centering
    \includegraphics[width=0.48\textwidth]{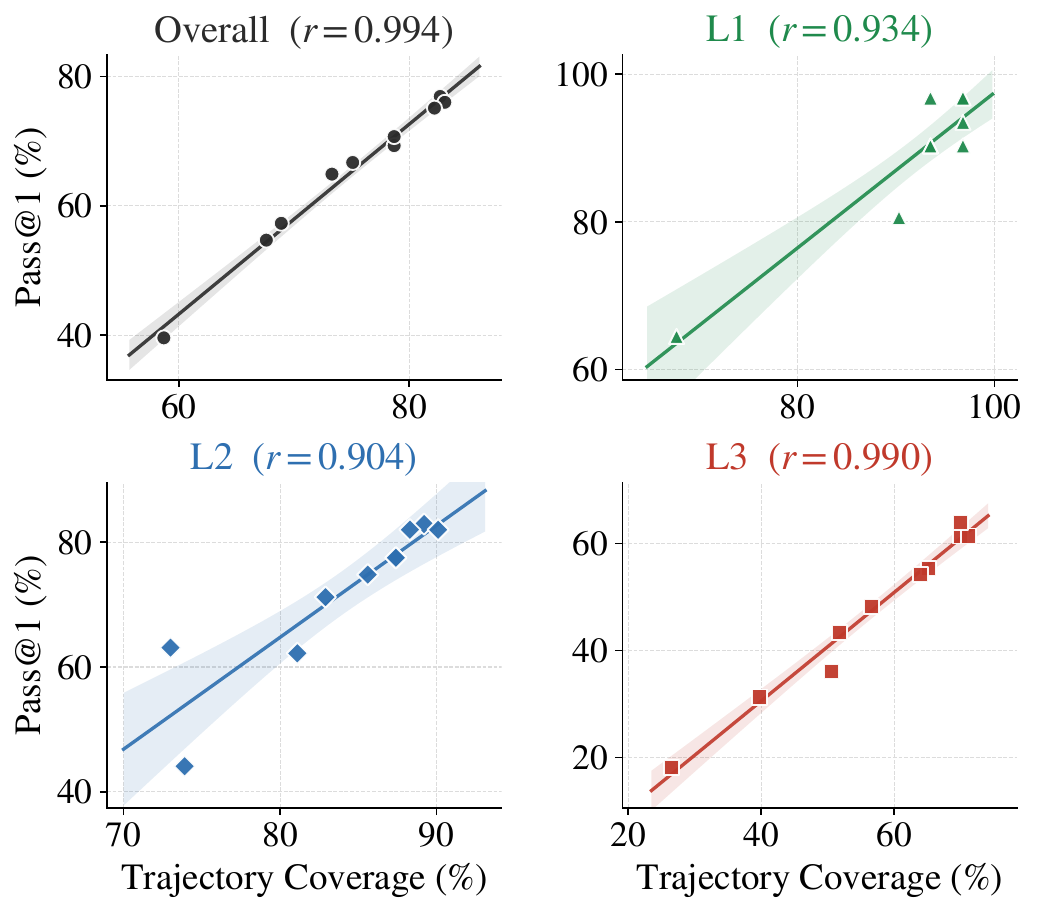}
    \caption{Correlation between trajectory coverage ($x$-axis) and Pass@1 ($y$-axis) across the ten evaluated models, with one panel per tier (Overall, L1, L2, L3). Each point is a model; the line shows a linear fit with its confidence band. The two metrics are strongly correlated at every tier ($r=0.994$ overall; $r=0.934/0.904/0.990$ for L1/L2/L3), supporting trajectory coverage as a meaningful proxy of agent evaluation quality.}
    \label{fig:pass1_trajcorr}
\end{figure}

\textbf{Effectiveness of trajectory-based evaluation. }Figure~\ref{fig:pass1_trajcorr} validates our trajectory signal: models with higher trajectory coverage achieve higher Pass@1 across all tiers, and the correlation remains especially strong on L3, where success relies on maintaining correct multi-step tool dependencies. This shows that tracking whether an agent follows mandatory tool calls is predictive of end-to-end success.

\textbf{Mismatch between coverage and Pass@3.} A notable exception arises for DeepSeek-V4. Despite relatively low Pass@1 on L3 (55.4\%), its Pass@3 on L3 reaches 73.5\%, the highest among all models. This suggests its primary limitation is not the absence of capability, but instability in single-trajectory long-horizon planning. With multiple samples, DeepSeek-V4 benefits from exploratory diversity, discovering alternative yet valid tool-calling routes even without strictly following the reference trajectory. These results expose a key trade-off: reasoning competence may be present but unreliable, manifesting only under repeated attempts that compensate for the low success probability of any single deterministic execution.

\textbf{Trajectory length.} The rightmost column of Table~\ref{tab:model_performance} reports the average trajectory length (number of interaction steps) for each model. Most models complete tasks within $3.25$--$4.15$ steps, with Gemini-3.1-Pro and DeepSeek-V4 producing the most concise trajectories ($3.25$). Longer trajectories do not yield better performance: MiniMax-M2.7 has the longest average trajectory ($4.15$) yet does not outperform more efficient models, suggesting that extra steps often reflect redundant retries or error recovery rather than productive exploration. In contrast, top-performing models execute shorter trajectories, indicating stronger planning that reduces unnecessary interactions.



\subsection{Error Analysis}
\label{sec:error_analysis}

We manually analyzed failed trajectories on \benchmarkname{} and found that errors in complex advertising analytics stem not from a single factor, but from the accumulation of multiple errors along the execution trajectory. Errors tend to cascade rather than occur in isolation: a single failed case can exhibit multiple error types, so their coverage rates need not sum to $100\%$. On average, each failed case contains about $1.06$--$1.27$ error types.
The error distribution is shown in Table~\ref{tab:error_distribution}.

\begin{table}[t]
\centering
\caption{Distribution of error types across models on \benchmarkname{}. We categorize each failure according to the root cause in the execution trace. TE. for Tool Error, PE. for Parameter Error, IF. for Instruction Following, CE. for Computation Error, Hal. for Hallucination, LC. for Long Context Ability. Statistics include all error types per case (multi-label). \textbf{Bold} for the best, while \underline{Underlined} for the second best performance.}
\label{tab:error_distribution}
\resizebox{0.48\textwidth}{!}{%
\begin{tabular}{lcccccccc}
\toprule
\multirow{2}{*}{\textbf{Model}} & \multicolumn{6}{c}{\textbf{Error Types}} & \multicolumn{2}{c}{\textbf{Total}} \\
\cmidrule(lr){2-7}\cmidrule(lr){8-9}
& \textbf{TE.} & \textbf{PE.} & \textbf{IF.} & \textbf{CE.} & \textbf{Hal.} & \textbf{LC.} & \textbf{\#Err} & \textbf{TotE} \\
\midrule
        Claude-Opus-4.7 & \textbf{15} & 8 & \underline{15} & \textbf{10} & \underline{3} & 4 & \textbf{52} & \textbf{55} \\
        Gemini-3.5-Flash & \textbf{15} & 12 & \textbf{14} & \underline{13} & \underline{3} & \textbf{1} & \underline{54} & \underline{58} \\
        Gemini-3.1-Pro & \underline{20} & \textbf{3} & 21 & 16 & \underline{3} & 4 & 56 & 67 \\
        Kimi-K2.6 & 25 & \underline{5} & 24 & 15 & \textbf{2} & \underline{3} & 66 & 74 \\
        GPT-5.5 & 21 & 12 & 19 & 18 & 7 & 4 & 69 & 81 \\
        DeepSeek-V4 & 25 & 22 & 18 & 14 & 4 & 4 & 79 & 87 \\
        MiniMax-M2.7 & 32 & 18 & \underline{15} & 19 & \underline{3} & 4 & 75 & 91 \\
        HY-3.0 & 22 & 30 & 20 & 21 & 6 & 10 & 96 & 109 \\
        \midrule
        Qwen3-235B & 47 & 14 & 22 & 35 & 8 & 4 & 102 & 130 \\
        Qwen3-30B-A3B & 61 & 24 & \underline{15} & 48 & 12 & 7 & 136 & 167 \\
\bottomrule
\end{tabular}}
\end{table} 

\paragraph{Tool and parameter errors.}
Across models, the dominant sources of failure are tool and parameter errors, together appearing in roughly $40$--$60\%$ of failed cases, indicating that multi-step tool orchestration remains brittle. Tool errors subsume planning and dependency mistakes, including incorrect tool ordering, dependency violations (e.g., querying report data before resolving the required account scope), premature termination, unnecessary detours, or invoking irrelevant tools, all of which drive execution away from the intended solution path.

Parameter errors frequently co-occur with planning issues and amplify their impact. Even when the agent selects an appropriate tool, it often mis-specifies key parameters such as date ranges, account or campaign filters, attribution windows, and aggregation definitions. These deviations are subtle but critical: individual steps can appear locally plausible while still yielding answers that disagree with ground truth. Such errors persist even in strong proprietary models, underscoring that maintaining precise alignment over long execution chains remains an open challenge in dynamic marketing settings.

\paragraph{Instruction-following errors.}
A distinct failure mode is instruction following: the agent constructs a plausible tool trajectory yet violates explicit constraints stated in the query, such as the requested time range, account or campaign filter, ranking criterion, or output format. Unlike tool errors, these failures occur even when the execution path is otherwise valid, showing that a correct tool sequence does not guarantee faithful adherence to user intent.

\paragraph{Retrieval failures and hallucination.}
Hallucination becomes particularly salient in tasks that require combining platform knowledge with live data. A recurring failure mode is that the agent produces a confident answer from prior assumptions instead of following the intended workflow to retrieve supporting evidence. Once the initial premise is incorrect, subsequent reasoning may remain coherent yet factually ungrounded, leading to high-confidence but unverifiable conclusions. This pattern is more pronounced in smaller models, suggesting limitations in evidence seeking and robust integration of retrieved knowledge under long-horizon interactions.

\paragraph{Computation and post-processing errors.}
As task complexity grows, agents must post-process large volumes of intermediate outputs, often via programmatic transformations. For tasks requiring Python-based processing, we observe a clear drop in code reliability as business logic becomes more complex, including syntax/runtime failures, missing key fields, incorrect slicing, and aggregation mismatches. Stronger models mitigate but do not eliminate these issues, indicating that translating natural-language intent into correct executable computation remains a major reliability bottleneck.

\paragraph{Long-context degradation.}
Although tool feedback can in principle enable self-correction~\cite{asai2024selfrag}, in practice error recovery frequently triggers repeated retries and verbose traces, rapidly expanding the interaction history. As context grows, agents struggle to track and reuse intermediate variables and to accurately localize the true source of failure. This often creates a negative feedback loop in which early mistakes lengthen the trajectory, which further degrades state tracking and induces additional errors, ultimately reducing both success rate and execution efficiency.

\section{Related Work}
\label{sec:related-works}

QA benchmarks, which construct fact-based questions across multiple dimensions, are extensively employed to evaluate the response effectiveness of LLMs or Deep Research systems. 
By providing static questions, these benchmarks aim to achieve comprehensive coverage~\cite{mmlu,gpqa,agieval} or construct more complex problems~\cite{mmlu-pro,browsecomp-plus,hle} to enhance discriminability.
Beyond static QA, a growing line of agent benchmarks evaluates interactive tool use, spanning general assistants~\cite{mialon2023gaia}, function calling~\cite{bfcl}, conversational tool agents~\cite{yao2024taubench,tau2bench}, and profession-aligned real-world tasks~\cite{xbench2025}.
Moving beyond static environments and fixed knowledge corpora, \textbf{\benchmarkname{}} introduces a suite of tools for accessing real-time accounts. 
It facilitates QA within a live sandbox environment that mirrors real-world advertising and marketing interactions, thereby enhancing the robustness of evaluations. 
Driven by a similar rationale, benchmarks like Mind2Web~\cite{mind2web} and WebwalkerQA~\cite{webwalker} also utilize unpredictable sandbox simulations to ensure real-time interactivity and environmental unpredictability.
\section{Conclusions}
\label{sec:conclusions}

We introduced \textbf{\benchmarkname{}}, a real-world, trajectory-aware advertising analytics benchmark for LLM agents. \benchmarkname{} couples execution trajectories with ground truths validated by human experts, enabling evaluation of both answer correctness and trajectory coverage. Experiments across a diverse set of models show strong performance stratification by difficulty and reveal that difficult queries remain a major bottleneck for current LLM agents.

\newpage
\section*{Limitations}

\paragraph{Overlap between the judge and the evaluated models.}
Gemini-3.1-Pro is both an evaluated agent and the automatic judge, which risks self-preference bias; this is limited by rubric-based judging against an explicit reference trajectory rather than open-ended scoring.

\paragraph{Imbalanced distribution across difficulty tiers.}
Sampled from real long-tail traffic, \benchmarkname{} has uneven tiers (L1$=$31, L2$=$111, L3$=$83), so overall scores favor the larger tiers and the small L1 tier gives limited resolution---close rankings there are effectively ties.

\section*{Ethical Statement}

Our research strictly adheres to ethical guidelines to safeguard user rights and privacy. 
All data within the evaluation benchmark has been anonymized and is used solely for scientific research purposes, aiming to advance more reliable evaluations in deep research systems.
\bibliography{custom}

\begin{thebibliography}{34}
\providecommand{\natexlab}[1]{#1}

\bibitem[{{Anthropic}(2026)}]{anthropic2026opus47}
{Anthropic}. 2026.
\newblock \href {https://www.anthropic.com/claude-opus-4-7-system-card} {Claude
  opus 4.7 system card}.
\newblock Closed-source large language model.

\bibitem[{Asai et~al.(2023)Asai, Wu, Wang, Sil, and
  Hajishirzi}]{asai2024selfrag}
Akari Asai, Zeqiu Wu, Yizhong Wang, Avirup Sil, and Hannaneh Hajishirzi. 2023.
\newblock Self-{RAG}: Learning to retrieve, generate, and critique through
  self-reflection.
\newblock In \emph{The Twelfth International Conference on Learning
  Representations}.

\bibitem[{Barres et~al.(2025)Barres, Dong, Ray, Si, and Narasimhan}]{tau2bench}
Victor Barres, Honghua Dong, Soham Ray, Xujie Si, and Karthik Narasimhan. 2025.
\newblock $\tau^2$-bench: Evaluating conversational agents in a dual-control
  environment.
\newblock \emph{arXiv preprint arXiv:2506.07982}.

\bibitem[{Chen et~al.(2025{\natexlab{a}})Chen, Ren, Liu, Hu, Tian, Xie, Liu,
  Zhang, Liu, Gong et~al.}]{xbench2025}
Kaiyuan Chen, Yixin Ren, Yang Liu, Xiaobo Hu, Haotong Tian, Tianbao Xie, Fangfu
  Liu, Haoye Zhang, Hongzhang Liu, Yuan Gong, and 1 others. 2025{\natexlab{a}}.
\newblock xbench: Tracking agents productivity scaling with profession-aligned
  real-world evaluations.
\newblock \emph{arXiv preprint arXiv:2506.13651}.

\bibitem[{Chen et~al.(2025{\natexlab{b}})Chen, Ma, Zhuang, Nie, Zou, Liu,
  Green, Patel, Meng, Su et~al.}]{browsecomp-plus}
Zijian Chen, Xueguang Ma, Shengyao Zhuang, Ping Nie, Kai Zou, Andrew Liu,
  Joshua Green, Kshama Patel, Ruoxi Meng, Mingyi Su, and 1 others.
  2025{\natexlab{b}}.
\newblock Browsecomp-plus: A more fair and transparent evaluation benchmark of
  deep-research agent.
\newblock \emph{arXiv preprint arXiv:2508.06600}.

\bibitem[{{DeepSeek-AI}(2026)}]{deepseek2026v4}
{DeepSeek-AI}. 2026.
\newblock \href {https://huggingface.co/deepseek-ai/DeepSeek-V4-Pro}
  {Deepseek-v4 technical report}.
\newblock Open-weights MoE model, MIT license.

\bibitem[{Deng et~al.(2023)Deng, Gu, Zheng, Chen, Stevens, Wang, Sun, and
  Su}]{mind2web}
Xiang Deng, Yu~Gu, Boyuan Zheng, Shijie Chen, Sam Stevens, Boshi Wang, Huan
  Sun, and Yu~Su. 2023.
\newblock Mind2web: Towards a generalist agent for the web.
\newblock \emph{Advances in Neural Information Processing Systems},
  36:28091--28114.

\bibitem[{{Google DeepMind}(2026{\natexlab{a}})}]{google2026gemini31pro}
{Google DeepMind}. 2026{\natexlab{a}}.
\newblock \href {https://deepmind.google/models/model-cards/gemini-3-1-pro/}
  {Gemini 3.1 pro model card}.
\newblock Closed-source (API only).

\bibitem[{{Google DeepMind}(2026{\natexlab{b}})}]{google2026gemini35flash}
{Google DeepMind}. 2026{\natexlab{b}}.
\newblock \href {https://deepmind.google/models/model-cards/gemini-3-5-flash/}
  {Gemini 3.5 flash model card}.
\newblock Closed-source (API only).

\bibitem[{Hendrycks et~al.(2020)Hendrycks, Burns, Basart, Zou, Mazeika, Song,
  and Steinhardt}]{mmlu}
Dan Hendrycks, Collin Burns, Steven Basart, Andy Zou, Mantas Mazeika, Dawn
  Song, and Jacob Steinhardt. 2020.
\newblock Measuring massive multitask language understanding.
\newblock \emph{arXiv preprint arXiv:2009.03300}.

\bibitem[{Lewis et~al.(2020)Lewis, Perez, Piktus, Petroni, Karpukhin, Goyal,
  K{\"u}ttler, Lewis, Yih, Rockt{\"a}schel et~al.}]{lewis2020rag}
Patrick Lewis, Ethan Perez, Aleksandra Piktus, Fabio Petroni, Vladimir
  Karpukhin, Naman Goyal, Heinrich K{\"u}ttler, Mike Lewis, Wen-tau Yih, Tim
  Rockt{\"a}schel, and 1 others. 2020.
\newblock Retrieval-augmented generation for knowledge-intensive nlp tasks.
\newblock \emph{Advances in neural information processing systems},
  33:9459--9474.

\bibitem[{Li et~al.(2026)}]{skillsbench2026}
Xiangyi Li and 1 others. 2026.
\newblock Skillsbench: Benchmarking how well agent skills work across diverse
  tasks.
\newblock \emph{arXiv preprint arXiv:2602.12670}.

\bibitem[{Mialon et~al.(2023)Mialon, Fourrier, Wolf, LeCun, and
  Scialom}]{mialon2023gaia}
Gr{\'e}goire Mialon, Cl{\'e}mentine Fourrier, Thomas Wolf, Yann LeCun, and
  Thomas Scialom. 2023.
\newblock Gaia: a benchmark for general ai assistants.
\newblock In \emph{The Twelfth International Conference on Learning
  Representations}.

\bibitem[{{MiniMax}(2026)}]{minimax2026m27}
{MiniMax}. 2026.
\newblock \href {https://huggingface.co/MiniMaxAI/MiniMax-M2.7} {Minimax-m2.7
  model card}.
\newblock Conditionally open-weights (Modified MIT).

\bibitem[{{OpenAI}(2026)}]{openai2026gpt55}
{OpenAI}. 2026.
\newblock \href {https://openai.com/index/introducing-gpt-5-5/} {Gpt-5.5}.
\newblock Closed-source large language model, released April 2026.

\bibitem[{Patil et~al.(2025)Patil, Mao, Ji, Yan, Suresh, Stoica, and
  Gonzalez}]{bfcl}
Shishir~G. Patil, Huanzhi Mao, Charlie Cheng-Jie Ji, Fanjia Yan, Vishnu Suresh,
  Ion Stoica, and Joseph~E. Gonzalez. 2025.
\newblock The berkeley function calling leaderboard (bfcl): From tool use to
  agentic evaluation of large language models.
\newblock In \emph{Forty-second International Conference on Machine Learning
  (ICML)}.

\bibitem[{Phan et~al.(2025)Phan, Gatti, Han, Li, Hu, Zhang, Zhang, Shaaban,
  Ling, Shi et~al.}]{hle}
Long Phan, Alice Gatti, Ziwen Han, Nathaniel Li, Josephina Hu, Hugh Zhang, Chen
  Bo~Calvin Zhang, Mohamed Shaaban, John Ling, Sean Shi, and 1 others. 2025.
\newblock Humanity's last exam.
\newblock \emph{arXiv preprint arXiv:2501.14249}.

\bibitem[{Rein et~al.(2024)Rein, Hou, Stickland, Petty, Pang, Dirani, Michael,
  and Bowman}]{gpqa}
David Rein, Betty~Li Hou, Asa~Cooper Stickland, Jackson Petty, Richard~Yuanzhe
  Pang, Julien Dirani, Julian Michael, and Samuel~R Bowman. 2024.
\newblock Gpqa: A graduate-level google-proof q\&a benchmark.
\newblock In \emph{First conference on language modeling}.

\bibitem[{Team et~al.(2025)Team, Bai, Bao, Chen, Chen, Chen, Chen, Chen, Chen,
  Chen et~al.}]{kimi-k2}
Kimi Team, Yifan Bai, Yiping Bao, Guanduo Chen, Jiahao Chen, Ningxin Chen,
  Ruijue Chen, Yanru Chen, Yuankun Chen, Yutian Chen, and 1 others. 2025.
\newblock Kimi k2: Open agentic intelligence.
\newblock \emph{arXiv preprint arXiv:2507.20534}.

\bibitem[{{Tencent Hunyuan Team}(2026)}]{tencent2026hy3}
{Tencent Hunyuan Team}. 2026.
\newblock \href {https://github.com/Tencent-Hunyuan/Hy3-preview} {Tencent
  hunyuan hy3 preview}.
\newblock Open-weights MoE (295B total params), Tencent Hunyuan Community
  License.

\bibitem[{Wang et~al.(2024{\natexlab{a}})Wang, Ma, Feng, Zhang, Yang, Zhang,
  Chen, Tang, Chen, Lin et~al.}]{wang2024llmagentsurvey}
Lei Wang, Chen Ma, Xueyang Feng, Zeyu Zhang, Hao Yang, Jingsen Zhang, Zhiyuan
  Chen, Jiakai Tang, Xu~Chen, Yankai Lin, and 1 others. 2024{\natexlab{a}}.
\newblock A survey on large language model based autonomous agents.
\newblock \emph{Frontiers of Computer Science}, 18(6):186345.

\bibitem[{Wang et~al.(2024{\natexlab{b}})Wang, Ma, Zhang, Ni, Chandra, Guo,
  Ren, Arulraj, He, Jiang et~al.}]{mmlu-pro}
Yubo Wang, Xueguang Ma, Ge~Zhang, Yuansheng Ni, Abhranil Chandra, Shiguang Guo,
  Weiming Ren, Aaran Arulraj, Xuan He, Ziyan Jiang, and 1 others.
  2024{\natexlab{b}}.
\newblock Mmlu-pro: A more robust and challenging multi-task language
  understanding benchmark.
\newblock \emph{Advances in Neural Information Processing Systems},
  37:95266--95290.

\bibitem[{Wei et~al.(2025)Wei, Sun, Papay, McKinney, Han, Fulford, Chung,
  Passos, Fedus, and Glaese}]{browsecomp}
Jason Wei, Zhiqing Sun, Spencer Papay, Scott McKinney, Jeffrey Han, Isa
  Fulford, Hyung~Won Chung, Alex~Tachard Passos, William Fedus, and Amelia
  Glaese. 2025.
\newblock Browsecomp: A simple yet challenging benchmark for browsing agents.
\newblock \emph{arXiv preprint arXiv:2504.12516}.

\bibitem[{Wong et~al.(2025)Wong, Wang, Zhao, Chen, Gao, Zhang, Zhou, Wang,
  Xiang, Zhang et~al.}]{wong2025widesearch}
Ryan Wong, Jiawei Wang, Junjie Zhao, Li~Chen, Yan Gao, Long Zhang, Xuan Zhou,
  Zuo Wang, Kai Xiang, Ge~Zhang, and 1 others. 2025.
\newblock Widesearch: Benchmarking agentic broad info-seeking.
\newblock \emph{arXiv preprint arXiv:2508.07999}.

\bibitem[{Wu et~al.(2025)Wu, Yin, Jiang, Wang, Xi, Fang, Zhang, He, Zhou, Xie
  et~al.}]{webwalker}
Jialong Wu, Wenbiao Yin, Yong Jiang, Zhenglin Wang, Zekun Xi, Runnan Fang,
  Linhai Zhang, Yulan He, Deyu Zhou, Pengjun Xie, and 1 others. 2025.
\newblock Webwalker: Benchmarking llms in web traversal.
\newblock In \emph{Proceedings of the 63rd Annual Meeting of the Association
  for Computational Linguistics (Volume 1: Long Papers)}, pages 10290--10305.

\bibitem[{Xi et~al.(2025)Xi, Chen, Guo, He, Ding, Hong, Zhang, Wang, Jin, Zhou
  et~al.}]{xi2025llmagentsurvey}
Zhiheng Xi, Wenxiang Chen, Xin Guo, Wei He, Yiwen Ding, Boyang Hong, Ming
  Zhang, Junzhe Wang, Senjie Jin, Enyu Zhou, and 1 others. 2025.
\newblock The rise and potential of large language model based agents: A
  survey.
\newblock \emph{Science China Information Sciences}, 68(2):121101.

\bibitem[{Xia et~al.(2025)Xia, Ding, Wan, Zhan, Du, and Tao}]{xia2025lact}
Tianle Xia, Liang Ding, Guojia Wan, Yibing Zhan, Bo~Du, and Dacheng Tao. 2025.
\newblock Improving complex reasoning over knowledge graph with logic-aware
  curriculum tuning.
\newblock In \emph{Proceedings of the AAAI Conference on Artificial
  Intelligence}, volume~39, pages 12881--12889.

\bibitem[{Xia et~al.(2026)}]{grasp2026}
Tianle Xia and 1 others. 2026.
\newblock Grasp: Graph-structured skill compositions for llm agents.
\newblock \emph{arXiv preprint arXiv:2604.17870}.

\bibitem[{Yang et~al.(2025)Yang, Li, Yang, Zhang, Hui, Zheng, Yu, Gao, Huang,
  Lv et~al.}]{qwen3}
An~Yang, Anfeng Li, Baosong Yang, Beichen Zhang, Binyuan Hui, Bo~Zheng, Bowen
  Yu, Chang Gao, Chengen Huang, Chenxu Lv, and 1 others. 2025.
\newblock Qwen3 technical report.
\newblock \emph{arXiv preprint arXiv:2505.09388}.

\bibitem[{Yao et~al.(2024)Yao, Shinn, Razavi, and Narasimhan}]{yao2024taubench}
Shunyu Yao, Noah Shinn, Pedram Razavi, and Karthik Narasimhan. 2024.
\newblock $\tau$-bench: A benchmark for tool-agent-user interaction in
  real-world domains.
\newblock \emph{arXiv preprint arXiv:2406.12045}.

\bibitem[{Yao et~al.(2022)Yao, Zhao, Yu, Du, Shafran, Narasimhan, and
  Cao}]{react}
Shunyu Yao, Jeffrey Zhao, Dian Yu, Nan Du, Izhak Shafran, Karthik~R Narasimhan,
  and Yuan Cao. 2022.
\newblock React: Synergizing reasoning and acting in language models.
\newblock In \emph{The eleventh international conference on learning
  representations}.

\bibitem[{Zeng et~al.(2025)Zeng, Liu, Chen, He, Liao, Tian, Wang, Wang, Yang,
  Yin et~al.}]{zeng2025futurex}
Zhiyuan Zeng, Jiashuo Liu, Siyuan Chen, Tianci He, Yali Liao, Yixiao Tian,
  Jinpeng Wang, Zaiyuan Wang, Yang Yang, Lingyue Yin, and 1 others. 2025.
\newblock Futurex: An advanced live benchmark for llm agents in future
  prediction.
\newblock \emph{arXiv preprint arXiv:2508.11987}.

\bibitem[{Zhong et~al.(2024)Zhong, Cui, Guo, Liang, Lu, Wang, Saied, Chen, and
  Duan}]{agieval}
Wanjun Zhong, Ruixiang Cui, Yiduo Guo, Yaobo Liang, Shuai Lu, Yanlin Wang, Amin
  Saied, Weizhu Chen, and Nan Duan. 2024.
\newblock Agieval: A human-centric benchmark for evaluating foundation models.
\newblock In \emph{Findings of the association for computational linguistics:
  NAACL 2024}, pages 2299--2314.

\bibitem[{Zhou et~al.(2023)Zhou, Xu, Zhu, Zhou, Lo, Sridhar, Cheng, Ou, Bisk,
  Fried et~al.}]{zhou2024webarena}
Shuyan Zhou, Frank~F Xu, Hao Zhu, Xuhui Zhou, Robert Lo, Abishek Sridhar,
  Xianyi Cheng, Tianyue Ou, Yonatan Bisk, Daniel Fried, and 1 others. 2023.
\newblock Webarena: A realistic web environment for building autonomous agents.
\newblock \emph{arXiv preprint arXiv:2307.13854}.

\end{thebibliography}

\newpage

\appendix

\section*{Appendix} 

\section{Prompt Templates}
\label{sec:appendix_prompts}

This appendix presents the complete prompt templates used in \benchmarkname{} for the agent system, report generation, and two-stage error analysis. Each prompt is shown verbatim (translated to English) to ensure full reproducibility.

\subsection{Agent System Prompt}
\label{sec:prompt_system}

The following prompt which defines the core system instruction for the advertising analytics agent is shown in Figure~\ref{fig:prompt_system}, including tool descriptions and output constraints.


\subsection{Report Generation Prompt}
\label{sec:prompt_report}

This prompt shown in Figure~\ref{fig:prompt_report} instructs the agent to generate a structured analytical report given data context and the user's query.


\subsection{Error Analysis: Stage 1 Prompt}
\label{sec:prompt_error_s1}

The Stage~1 prompt shown in Figure~\ref{fig:prompt_error_stage1_part1} and ~\ref{fig:prompt_error_stage1_part2} performs a rapid assessment of result correctness, trajectory coverage, instruction compliance, parameter errors, redundant calls, and dependency violations.





\section{Real-World Deployment: Benchmark-Guided Iteration}
\label{sec:deployment}

Beyond one-off offline evaluation, \benchmarkname{} is deployed as a continuous evaluation harness for a production advertising-analytics agent. Each candidate version is scored on \benchmarkname{}, and its trajectory-level diagnostics localize concrete failure modes that directly inform the next iteration. We track real-world impact with an online user satisfaction rate---the fraction of thumbs-up collected from a crowd-sourced evaluation. As shown in Table~\ref{tab:deployment}, benchmark-guided iteration steadily improves this metric from $65.55\%$ to $88.37\%$ over three versions. Crucially, the online satisfaction rate moves in lockstep with the agent's overall accuracy on \benchmarkname{}, which rises from $57.62\%$ to $70.22\%$ over the same versions, confirming that gains diagnosed on \benchmarkname{} translate into a better user experience in production.

\begin{table}[t]
\centering
\small
\begin{tabular}{lccc}
\toprule
\textbf{Metric (\%)} & \textbf{V1} & \textbf{V2} & \textbf{V3} \\
\midrule
\benchmarkname{} Pass@1 & 57.62 & 63.99 & \textbf{70.22} \\
Online user satisfaction & 65.55 & 84.45 & \textbf{88.37} \\
\bottomrule
\end{tabular}
\caption{Benchmark-guided iteration of a production advertising-analytics agent across successive versions. Improvements on \benchmarkname{} track closely with the online user satisfaction rate (thumbs-up ratio from a crowd-sourced evaluation), which rises to $88.37\%$.}
\label{tab:deployment}
\end{table}

\section{Tool Inventory}
\label{sec:tool_inventory}

The advertising analytics agent is equipped with nine tools shown in Table~\ref{tab:tool_list}.

\noindent Figure~\ref{fig:tool_dep} shows a typical tool invocation tree. The agent begins with the user query, then issues parallel calls to retrieve domain knowledge, resolve account scope, and fetch peer benchmarks (Round~2). Subsequent rounds depend on earlier results: ad-level configuration requires account resolution (Round~3), data retrieval requires ad context (Round~4), computation integrates all prior outputs (Round~5), and the final report is generated (Round~6).

\section{Case Studies Across Difficulty Levels}
\label{sec:case_study}

\begin{figure*}[htbp]
\centering
\includegraphics[width=0.95\textwidth]{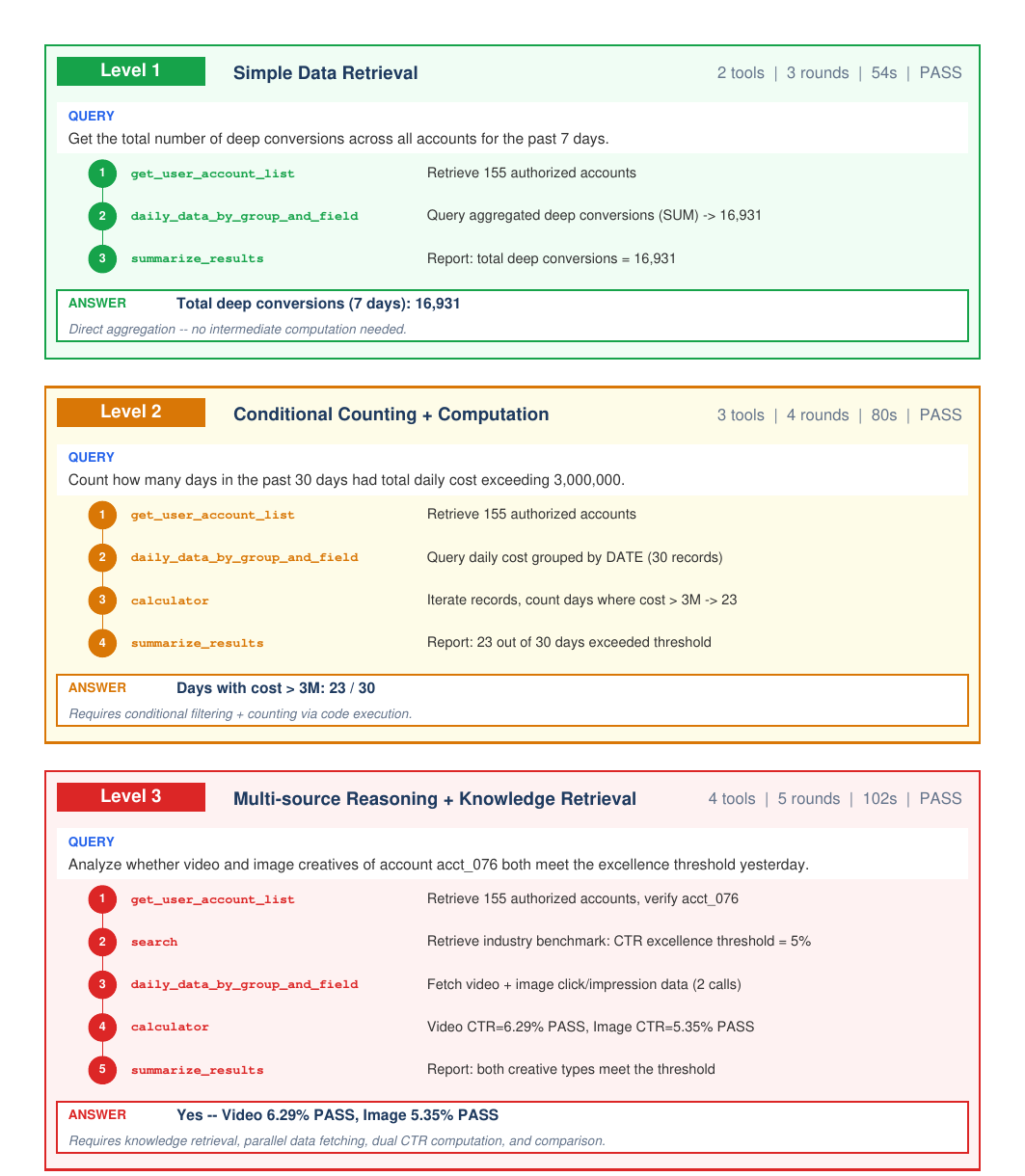}
\caption{Overview comparison of three representative cases across difficulty levels. L1 requires direct retrieval; L2 adds conditional computation; L3 demands knowledge retrieval, parallel data fetching, and dual-metric verification.}
\label{fig:case_overview}
\end{figure*}

To illustrate the progressive complexity of \benchmarkname{}, we present three representative cases solved by Gemini-3.1-Pro, one from each difficulty level. All account IDs and user IDs are anonymized. Figure~\ref{fig:case_overview} provides a side-by-side comparison of the three cases, highlighting the escalating demands in tool usage and reasoning depth.

\section{Error Analysis Examples}
\label{sec:error_analysis_examples}

\begin{figure*}[htbp]
\centering
\includegraphics[width=0.95\textwidth, page=1]{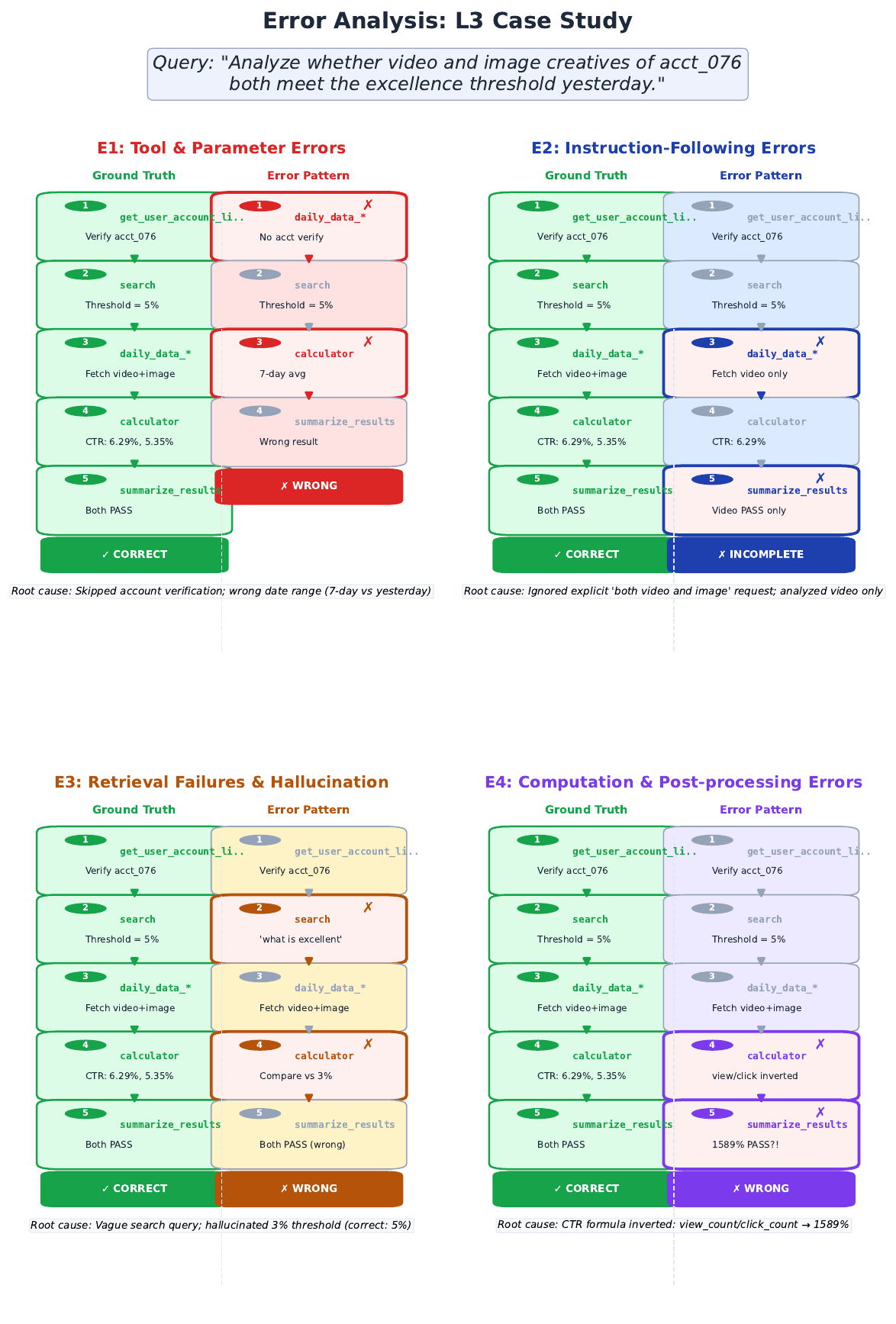}
\caption{Error analysis of the L3 case. Each subplot pairs the correct trajectory (left) with one error pattern (right). Error steps are marked with \textcolor{red}{\ding{55}}. E1: tool \& parameter; E2: instruction-following; E3: retrieval \& hallucination; E4: computation \& post-processing.}
\label{fig:error_analysis}
\end{figure*}

Using the L3 case as a reference, Figure~\ref{fig:error_analysis} illustrates four representative error patterns observed during LLM agent evaluation. Each subplot pairs the ground-truth trajectory (left, green) with an erroneous trajectory (right), with error steps highlighted: \textbf{E1}~tool \& parameter errors, \textbf{E2}~instruction-following errors, \textbf{E3}~retrieval failure \& hallucination, and \textbf{E4}~computation \& post-processing errors.


\begin{figure}[htbp]
\centering
\resizebox{0.48\textwidth}{!}{%
\begin{tikzpicture}[
    >=Stealth,
    node distance=1.1cm and 0.6cm,
    every node/.style={font=\small},
    box/.style={
        draw=black!70, rounded corners=2pt, minimum height=0.7cm,
        minimum width=2.2cm, align=center, inner sep=3pt,
        font=\small\ttfamily, fill=black!2
    },
    querybox/.style={
        draw=black!80, rounded corners=3pt, minimum height=0.75cm,
        minimum width=2.6cm, align=center, inner sep=4pt,
        font=\small\bfseries, fill=black!4
    },
    arr/.style={->, thick, color=black!50},
    darr/.style={->, densely dashed, thin, color=black!30},
    rlabel/.style={font=\scriptsize\bfseries, color=black!50},
    sublabel/.style={font=\scriptsize, color=black!45, below=0pt},
]
\node[querybox] (query) {User Query};
\node[sublabel] at (query.south) {natural-language question};
\node[rlabel, left=0.3cm of query] {R1};

\node[box, below left=1.1cm and 2.0cm of query] (search) {search};
\node[sublabel] at (search.south) {domain knowledge};

\node[box, below=1.1cm of query] (acctlist) {get\_user\_account\_list};
\node[sublabel] at (acctlist.south) {resolve accounts};

\node[box, below right=1.1cm and 2.0cm of query] (creative) {get\_top\_good\_creative};
\node[sublabel] at (creative.south) {peer benchmarks};

\node[rlabel] at ($(search.west)+(-0.35,0)$) {R2};

\draw[arr] (query) -- (search);
\draw[arr] (query) -- (acctlist);
\draw[arr] (query) -- (creative);

\node[box, below=1.5cm of acctlist] (adgroup) {get\_account\_adgroup\_info};
\node[sublabel] at (adgroup.south) {ad-level configuration};
\node[rlabel] at ($(adgroup.west)+(-0.35,0)$) {R3};

\draw[arr] (acctlist) -- (adgroup);

\node[font=\tiny, color=black!35, right=0.08cm] at ($(acctlist.south)!0.5!(adgroup.north)+(0.7,0)$) {\textit{requires account}};

\node[box, below left=1.5cm and 1.3cm of adgroup] (daily) {daily\_data\_*};
\node[sublabel] at (daily.south) {daily metrics};

\node[box, below right=1.5cm and 1.3cm of adgroup] (hourly) {hourly\_data\_*};
\node[sublabel] at (hourly.south) {hourly metrics};

\node[rlabel] at ($(daily.west)+(-0.35,0)$) {R4};

\draw[arr] (adgroup) -- (daily);
\draw[arr] (adgroup) -- (hourly);

\node[box, below=1.5cm of adgroup, yshift=-2.4cm] (calc) {calculator};
\node[sublabel] at (calc.south) {compute \& analyze};
\node[rlabel] at ($(calc.west)+(-0.35,0)$) {R5};

\draw[arr] (daily) -- (calc);
\draw[arr] (hourly) -- (calc);

\draw[darr] (search.south) .. controls +(0,-2.5) and +(-1.5,1) .. (calc.north west);
\draw[darr] (creative.south) .. controls +(0,-2.5) and +(1.5,1) .. (calc.north east);

\node[box, below=1.1cm of calc] (summary) {summarize\_results};
\node[sublabel] at (summary.south) {output report};
\node[rlabel] at ($(summary.west)+(-0.35,0)$) {R6};

\draw[arr] (calc) -- (summary);

\end{tikzpicture}%
}
\caption{A typical tool invocation tree. Solid arrows indicate direct dependencies; dashed arrows indicate indirect dependencies (domain knowledge and benchmarks feed into computation).}
\label{fig:tool_dep}
\end{figure}

\begin{figure*}[htbp]
\centering
\begin{tcolorbox}[colback=blue!3, colframe=blue!40, title=\textbf{Agent System Prompt}, fonttitle=\bfseries, boxrule=0.5pt, left=3mm, right=3mm, top=2mm, bottom=2mm]
You are an advertising marketing assistant, specializing in decomposing complex data query tasks into specific tool calls to retrieve data, providing users with precise marketing data.

\textbf{Core Task}
\begin{enumerate}[leftmargin=*, nosep]
\item Analyze the user's request and conversation history, plan the data needed at the current stage, and execute tool calls accordingly.
\end{enumerate}

\textbf{Output Requirements}
\begin{enumerate}[leftmargin=*, nosep]
\item Responses must include planning and tool calls. Output the following tags:\\
  \texttt{<plan>PLAN: </plan>} \quad What objective information the current step expects to collect.\\
  \texttt{<tools></tools>} \quad List of tool calls.
\item \texttt{<plan>} format:
\begin{verbatim}
<plan>
PLAN: Describe the current step goal and what information to collect.
</plan>
\end{verbatim}
\item \texttt{<tools>} format (strict):
\begin{verbatim}
<tools>
  <tool>
  TOOL_NAME: tool_name
  INPUT: tool input (JSON)
  </tool>
</tools>
\end{verbatim}
  Note: Each turn outputs one \texttt{<tools>} block. Nest multiple \texttt{<tool>} for parallel calls.
  Results appear in \texttt{<tool\_res>...</tool\_res>}.
\end{enumerate}

\textbf{Working Rules}
\begin{enumerate}[leftmargin=*, nosep]
\item Answer based on the user's account data, providing customized solutions.
\item \texttt{<plan>} specifies what information to collect; \texttt{<tools>} implements the plan.
\item Multiple queries can nest multiple \texttt{<tool>} blocks. `\#' comments are prohibited in INPUT.
\item For unavailable data/parameters, use \texttt{summarize\_results} to indicate inability to retrieve.
\item If no specific account is mentioned, analyze all accounts with permissions.
\item All authorized accounts must be obtained from \texttt{get\_user\_account\_list} with full \texttt{account\_id\_list}.
\item Calculator results must use \texttt{print()}.
\item ``Last week'' refers to last Monday through last Sunday.
\end{enumerate}

\textbf{Available Tools}
\begin{itemize}[leftmargin=*, nosep]
\item \texttt{search} -- Knowledge retrieval (\texttt{query}: str, max 20 chars)
\item \texttt{daily\_data\_by\_group\_and\_field} -- Daily-granularity report data\\
  (\texttt{user\_id}, \texttt{begin}, \texttt{end}, \texttt{group\_by\_type}, \texttt{fields}, \texttt{account\_id\_list}, \texttt{page\_size}, \ldots)
\item \texttt{hourly\_data\_by\_group\_and\_field} -- Hourly-granularity report for a specific date\\
  (\texttt{user\_id}, \texttt{date}, \texttt{group\_by\_type}, \texttt{fields}, \texttt{page\_size}, \ldots)
\item \texttt{calculator} -- Execute business computation code (\texttt{code}: str)
\item \texttt{get\_account\_info} -- Account basic info (\texttt{user\_id}, \texttt{account\_id\_list})
\item \texttt{get\_account\_adgroup\_info} -- Ad group info (\texttt{user\_id}, \texttt{account\_id}, \texttt{adgroup\_id\_list})
\item \texttt{get\_user\_account\_list} -- List of authorized accounts (\texttt{user\_id}, \texttt{page\_size}, \ldots)
\item \texttt{summarize\_results} -- Summarize results (\texttt{query}: ``Resolved'' $|$ ``Unresolved'')
\item \texttt{get\_top\_good\_creative} -- Top peer creatives (\texttt{industry}, \texttt{site\_set}, \texttt{creative\_type})
\end{itemize}

\end{tcolorbox}
\caption{Agent system prompt for the advertising marketing assistant. The agent decomposes user queries into structured planning and tool-call sequences.}
\label{fig:prompt_system}
\end{figure*}

\begin{figure*}[htbp]
\centering
\begin{tcolorbox}[colback=blue!3, colframe=blue!40, title=\textbf{Report Generation Prompt}, fonttitle=\bfseries, boxrule=0.5pt, left=3mm, right=3mm, top=2mm, bottom=2mm]
You are an advertising data analyst. Based on the user's question and the data context, summarize the situation and provide a direct, precise answer to the data metrics the user cares about, strictly following the user's requirements.

\textbf{Output Requirements}
\begin{enumerate}[leftmargin=*, nosep]
\item Output data directly without scientific notation or comma-separated formats.
\item Round to two decimal places (e.g., 0.11 and 0.1091 are the same).
\item Analyze and answer based on the user's question, not indiscriminate summarization.
\item For knowledge-based Q\&A, provide a brief explanation of 10--20 words.
\end{enumerate}

\textbf{Input}

\textbf{User Question:} \texttt{\{query\}}

\textbf{Data Context:}

\texttt{<context>}\\
\texttt{\{context\}}\\
\texttt{</context>}

\end{tcolorbox}
\caption{Prompt template for report generation. The LLM analyst synthesizes tool-returned data into a precise, user-facing answer.}
\label{fig:prompt_report}
\end{figure*}

\begin{table*}[htbp]
\centering
\small
\resizebox{0.95\textwidth}{!}{%
\begin{tabular}{l p{8cm}}
\toprule
\textbf{Tool} & \textbf{Description} \\
\midrule
\texttt{get\_user\_account\_list} & Retrieve the list of authorized advertising accounts for a given user, including account ID, company name, industry, and last consumption date. \\
\texttt{get\_account\_info} & Fetch account metadata: company name, industry, daily budget, audit status, and audit reason. \\
\texttt{get\_account\_adgroup\_info} & Fetch ad-level configuration: ad name, status, schedule, targeting, bid, placement, and marketing objective. \\
\texttt{daily\_data\_by\_group\_and\_field} & Fetch daily-granularity report data (cost, clicks, impressions, CTR, conversions, etc.) with flexible grouping dimensions (date, account, ad, creative, region, etc.). \\
\texttt{hourly\_data\_by\_group\_and\_field} & Fetch hourly-granularity report data for a single specified day, supporting grouping by hour, ad, or creative. \\
\texttt{search} & Retrieve domain knowledge, best practices, and industry benchmarks via keyword query. \\
\texttt{calculator} & Execute Python code snippets for arithmetic and statistical computation on retrieved data. \\
\texttt{get\_top\_good\_creative} & Fetch top-performing peer creatives from the past 3 days by industry, placement, and creative type. \\
\texttt{summarize\_results} & Terminate the session and summarize findings; mark the query as resolved or unresolved. \\
\bottomrule
\end{tabular}
}
\caption{Tool inventory of the advertising analytics agent.}
\label{tab:tool_list}
\end{table*}

\begin{figure*}[htbp]
\centering
\begin{tcolorbox}[colback=blue!3, colframe=blue!40, title=\textbf{Error Analysis Stage 1: Correctness and Coverage Prompt}, fonttitle=\bfseries, boxrule=0.5pt, left=3mm, right=3mm, top=2mm, bottom=2mm]
Evaluate the execution quality of the following advertising analytics case.

\textbf{Available Tools and Dependencies}

\textbf{Independent Tools} (no dependencies, can be called directly)

\begin{enumerate}[leftmargin=*, nosep]
\item \texttt{search} -- Retrieve advertising/marketing domain knowledge
  \begin{itemize}[leftmargin=*, nosep]
  \item Parameters: \texttt{query} (required, max 20 chars)
  \item Dependencies: None
  \end{itemize}

\item \texttt{get\_top\_good\_creative} -- Get top-performing creatives from the platform
  \begin{itemize}[leftmargin=*, nosep]
  \item Parameters: \texttt{query} (filter description: industry/placement/creative type/keywords)
  \item Dependencies: None
  \end{itemize}

\item \texttt{get\_user\_account\_list} -- Get all account IDs under the user
  \begin{itemize}[leftmargin=*, nosep]
  \item Parameters: \texttt{user\_id} (required), \texttt{page\_size}
  \item Returns: Account ID list
  \item Dependencies: None (but is a prerequisite for data query tools)
  \end{itemize}
\end{enumerate}

\textbf{Data Query Tools} (require \texttt{account\_id})

\begin{enumerate}[leftmargin=*, nosep, start=4]
\item \texttt{daily\_data\_by\_group\_and\_field} -- Daily-granularity report data
  \begin{itemize}[leftmargin=*, nosep]
  \item Parameters: \texttt{user\_id}, \texttt{begin}, \texttt{end}, \texttt{group\_by\_type}, \texttt{fields}, \texttt{account\_id\_list}, \texttt{page\_size}, \texttt{adgroup\_id} (opt.), \texttt{order\_by} (opt.)
  \item \texttt{group\_by\_type}: SUM / DATE / MONTH / WEEK / SITE\_SET / ADGROUP\_ID / ACCOUNT\_ID / CREATIVE\_ID / MATERIAL\_VIDEO / MATERIAL\_IMAGE / MARKETING\_ASSET / GENDER / AGE / REGION / CITY
  \item \texttt{fields}: cost / valid\_click\_count / view\_count / ctr / cpc / conversions\_cost / conversions\_count / conversions\_rate / deep\_conversions\_count / deep\_conversions\_rate / deep\_conversions\_cost
  \item Dependencies: Requires \texttt{account\_id\_list} (from \texttt{get\_user\_account\_list})
  \end{itemize}

\item \texttt{hourly\_data\_by\_group\_and\_field} -- Hourly-granularity report for a specific date
  \begin{itemize}[leftmargin=*, nosep]
  \item Parameters: \texttt{user\_id}, \texttt{date}, \texttt{group\_by\_type}, \texttt{fields}, \texttt{page\_size}, \texttt{account\_id\_list} (opt.), \texttt{adgroup\_id} (opt.), \texttt{order\_by} (opt.)
  \item \texttt{group\_by\_type}: SUM / HOUR / ACCOUNT\_ID / ADGROUP\_ID / ADGROUP\_ID\_AND\_HOUR / CREATIVE\_ID / CREATIVE\_ID\_AND\_HOUR
  \item Dependencies: Requires \texttt{account\_id\_list} (from \texttt{get\_user\_account\_list})
  \end{itemize}

\item \texttt{get\_account\_info} -- Get account basic information
  \begin{itemize}[leftmargin=*, nosep]
  \item Parameters: \texttt{user\_id}, \texttt{account\_id\_list}
  \item Returns: Account name, industry, daily budget, review status, etc.
  \item Dependencies: Requires \texttt{account\_id\_list}
  \end{itemize}

\item \texttt{get\_account\_adgroup\_info} -- Get ad group list / ad details
  \begin{itemize}[leftmargin=*, nosep]
  \item Parameters: \texttt{user\_id}, \texttt{account\_id}, \texttt{adgroup\_id\_list} (opt.)
  \item Returns: Ad name, status, dates, bid, budget, targeting, placement, etc.
  \item Dependencies: Requires \texttt{account\_id}
  \end{itemize}
\end{enumerate}

\textbf{Computation Tool}

\begin{enumerate}[leftmargin=*, nosep, start=8]
\item \texttt{calculator} -- Write Python code to compute business metrics
  \begin{itemize}[leftmargin=*, nosep]
  \item Parameters: \texttt{code} (required)
  \item Dependencies: Usually requires data to be fetched first
  \item Note: Return field names are like ``Cost (CNY)'', ``Click Count'', ``Target Conversions'', etc.
  \end{itemize}
\end{enumerate}

\end{tcolorbox}
\caption{Prompt template for Stage~1 error analysis (Part~1: Tool Definitions). Lists all available tools, their parameters, and dependency requirements.}
\label{fig:prompt_error_stage1_part1}
\end{figure*}

\begin{figure*}[htbp]
\centering
\begin{tcolorbox}[colback=blue!3, colframe=blue!40, title=\textbf{Error Analysis Stage 1: Correctness and Coverage Prompt (cont.)}, fonttitle=\bfseries, boxrule=0.5pt, left=3mm, right=3mm, top=2mm, bottom=2mm]

\textbf{Tool Dependency Graph}

\begin{verbatim}
search / get_top_good_creative (independent, callable anytime)

get_user_account_list (prerequisite for data queries)
  |-> daily_data_by_group_and_field (needs account_id_list)
  |-> hourly_data_by_group_and_field (needs account_id_list)
  |-> get_account_info (needs account_id_list)
  |-> get_account_adgroup_info (needs account_id)
         |-> daily_data/hourly_data (can query by adgroup_id)

After data retrieval -> calculator (perform computation)
\end{verbatim}

\textbf{Common Error Patterns}
\begin{itemize}[leftmargin=*, nosep]
\item Calling tools that require \texttt{account\_id} without first calling \texttt{get\_user\_account\_list}
\item Mixing up \texttt{daily\_data} and \texttt{hourly\_data} (different granularity scenarios)
\item Incorrect \texttt{group\_by\_type} (e.g., using SUM when aggregation by account is needed)
\item Redundant calls to \texttt{get\_user\_account\_list}
\item Wrong field names in calculator (e.g., using ``cost'' instead of ``Cost (CNY)'')
\end{itemize}

\medskip
\textbf{User Query:} \texttt{\{query\}}

\textbf{Ground-Truth Answer:} \texttt{\{ground\_truth\}}

\textbf{Expected Trajectory:} \texttt{\{expected\_trajectory\}}

\textbf{Actual Trajectory:} \texttt{\{trajectory\}}

\textbf{Model's Final Answer:} \texttt{\{final\_answer\}}

\medskip
Based on the tool definitions and dependency relationships, respond in the following JSON format only (no other content):

\begin{verbatim}
{
    "result_correct": true/false, "trajectory_covered": true/false,
    "has_instruction_violation": true/false, "has_error_parameters": true/false,
    "has_redundant_calls": true/false, "has_dependency_error": true/false,
    "brief_reason": ""
}
\end{verbatim}

\textbf{Field descriptions:}
\begin{itemize}[leftmargin=*, nosep]
\item \texttt{result\_correct}: Whether the model's answer matches the ground truth (format differences like thousands separators or units are acceptable)
\item \texttt{trajectory\_covered}: Whether the actual trajectory covers all necessary tool calls in the expected trajectory
\item \texttt{has\_instruction\_violation}: Whether there are turns with no tool calls (status ``no tool call'')
\item \texttt{has\_error\_parameters}: Whether parameters are incorrect (e.g., target is ad ID but parameter is filled with account ID; target is conversions but parameter is cost)
\item \texttt{has\_redundant\_calls}: Whether there are unnecessary repeated calls (previously called tools forgotten)
\item \texttt{has\_dependency\_error}: Whether tool dependency relationships are violated (e.g., calling tools that need \texttt{account\_id} without first obtaining it; calling calculator without fetching data first)
\item \texttt{brief\_reason}: Reasoning for the judgment (max 50 words)
\end{itemize}

\end{tcolorbox}
\caption{Prompt template for Stage~1 error analysis (Part~2: Dependency Graph, Error Patterns, and Evaluation Output). The LLM judge evaluates execution correctness, trajectory coverage, instruction compliance, parameter accuracy, redundancy, and dependency violations.}
\label{fig:prompt_error_stage1_part2}
\end{figure*}

\end{document}